\documentclass[journal]{IEEEtran}
\usepackage{amsmath,amsfonts,amssymb}
\usepackage{algorithmic}
\usepackage{algorithm}
\usepackage{array}
\usepackage[caption=false,font=normalsize,labelfont=sf,textfont=sf]{subfig}
\usepackage{textcomp}
\usepackage{stfloats}
\usepackage{url}
\usepackage{verbatim}
\usepackage{graphicx}
\usepackage{cite}
\usepackage{multirow}
\usepackage[table]{xcolor}

\usepackage{color}
\usepackage{booktabs}
\usepackage{makecell}

\usepackage[colorlinks,bookmarksopen,bookmarksnumbered,citecolor=green, linkcolor=red, urlcolor=magenta]{hyperref}

\begin{document}

\title{Preference-Guided Debiasing for No-Reference Enhancement Image Quality Assessment}

\author{
Shiqi Gao, Kang Fu, Zitong Xu, Huiyu Duan, Xiongkuo Min, Jia Wang, Guangtao Zhai\IEEEauthorrefmark{2}%
\thanks{\IEEEauthorrefmark{2} Co-corresponding authors.}%
\thanks{Shiqi Gao, Kang Fu, Zitong Xu, Huiyu Duan, Xiongkuo Min, Jia Wang and Guangtao Zhai are with the Institute of Image Communication and Network Engineering, Shanghai Jiao Tong University, Shanghai 200240, China (e-mail: gaoshiqi@sjtu.edu.cn; fuk20-20@sjtu.edu.cn; xuzitong@sjtu.edu.cn; huiyuduan@sjtu.edu.cn; minxiongkuo@sjtu.edu.cn; jiawang@sjtu.edu.cn; zhaiguangtao@sjtu.edu.cn).}%
}



\maketitle

\begin{abstract}
Current no-reference image quality assessment (NR-IQA) models for enhanced images often struggle to generalize, as they tend to overfit to the distinct patterns of specific enhancement algorithms rather than evaluating genuine perceptual quality.
To address this issue, we propose a preference-guided debiasing framework for no-reference enhancement image quality assessment (EIQA). Specifically, we first learn a continuous enhancement-preference embedding space using supervised contrastive learning, where images generated by similar enhancement styles are encouraged to have closer representations. Based on this, we further estimate the enhancement-induced nuisance component contained in the raw quality representation and remove it before quality regression. In this way, the model is guided to focus on algorithm-invariant perceptual quality cues instead of enhancement-specific visual fingerprints. To facilitate stable optimization, we adopt a two-stage training strategy that first learns the enhancement-preference space and then performs debiased quality prediction.
Extensive experiments on public EIQA benchmarks demonstrate that the proposed method effectively mitigates algorithm-induced representation bias and achieves superior robustness and cross-algorithm generalization compared with existing approaches.
\end{abstract}




\section{Introduction}

Image enhancement aims to improve the perceptual quality of images captured under challenging conditions~\cite{gao2024quality,duan2022develop,duan2024uniprocessor}, such as low-light environments, haze, or sensor noise. With the rapid development of learning-based enhancement algorithms~\cite{retinexnet, zero_dce, kinD}, evaluating the perceptual quality of enhanced images has become increasingly important. Enhancement image quality assessment (EIQA) seeks to predict human-perceived quality for enhanced images and plays an important role in benchmarking enhancement algorithms, guiding model development, and enabling quality-aware image processing systems~\cite{lieqa}.

Image quality assessment (IQA) has been extensively studied in the literature~\cite{ssim, BRISQUE}. Traditional no-reference IQA methods often rely on natural scene statistics or handcrafted features to estimate perceptual quality~\cite{BRISQUE, NIQE}, while more recent approaches learn quality-aware representations directly from images using deep neural networks~\cite{HyperIQA, MUSIQ}. Although these methods have achieved strong performance on conventional distortions such as compression artifacts, blur, and noise, their underlying assumptions do not readily transfer to EIQA~\cite{squarelol, nrlowlightqa}.

Compared with conventional IQA, EIQA presents a distinct challenge. Typical distortions mainly degrade image quality, whereas enhancement algorithms intentionally alter image appearance in an attempt to improve perceptual quality. Such alterations often involve complex changes in tone, contrast, texture, and artifact suppression. As a result, images generated by different enhancement algorithms may exhibit substantially different visual styles even when their perceptual quality is similar~\cite{squarelol, rlie_iqa}. This additional algorithm-dependent variation makes reliable quality modeling more difficult.

Existing EIQA methods typically learn to regress mean opinion scores (MOS) directly from enhanced images using convolutional or transformer-based networks~\cite{squarelol,lieqa,HyperIQA,rlie_iqa}. While effective in standard evaluation settings, these methods generally do not explicitly distinguish intrinsic perceptual quality cues from enhancement-style variation. When training data cover only a limited set of enhancement algorithms, the model may inadvertently rely on algorithm-specific visual patterns rather than genuinely quality-relevant evidence. Consequently, the learned representations can become biased toward enhancement style, leading to inferior generalization to images produced by previously unseen algorithms.

To address this issue, we propose to explicitly model \emph{enhancement preference}, i.e., the algorithm-related visual style tendency shared by images enhanced with similar methods, and use it to guide quality representation learning. Instead of formulating enhancement modeling as a closed-set algorithm classification problem, we learn a continuous embedding space that captures enhancement-style similarity via supervised contrastive learning. This preference embedding serves as a compact representation of algorithm-induced appearance characteristics.

\begin{figure}
\centering
\includegraphics[width=0.99\linewidth]{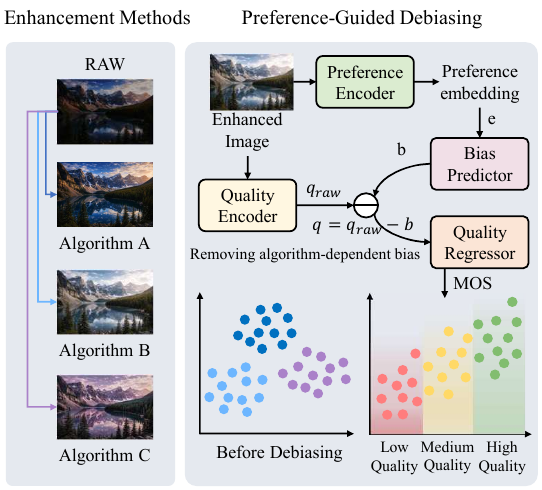}
\caption{Different enhancement algorithms may produce distinct visual styles, causing algorithm-dependent bias in quality representation. The proposed preference-guided debiasing framework explicitly models enhancement preference and removes its interference, leading to more reliable perceptual quality prediction.}
\label{fig1}
\end{figure}

Building on this representation, we introduce a preference-guided debiasing framework for EIQA. The key idea is to estimate the enhancement-induced nuisance component embedded in the raw quality representation and remove it before quality regression. By suppressing style-related bias in the feature space, the model is encouraged to focus on algorithm-invariant perceptual cues that are more directly related to human quality judgments. To stabilize optimization, we adopt a two-stage training strategy, \textit{i.e.}, we first learn a robust enhancement-preference embedding space, and then perform preference-guided quality prediction based on debiased features.
Extensive experiments demonstrate that the proposed method effectively mitigates algorithm-induced representation bias and yields superior robustness and cross-algorithm generalization in enhancement image quality assessment. The schematic diagram is shown in Fig.~\ref{fig1}.

The main contributions of this work are summarized as follows:
\begin{itemize}
\item We identify the algorithm-induced bias problem in enhancement image quality assessment and show that enhancement-specific visual styles can interfere with learning intrinsic perceptual quality representations.

\item We propose a preference-guided debiasing framework that explicitly models enhancement-style preference in a continuous embedding space and removes algorithm-induced nuisance components from quality features.

\item We develop a two-stage training strategy that improves the robustness of quality prediction and enhances generalization to images generated by unseen enhancement algorithms.
\end{itemize}

\section{Related Work}
\label{sec:related}

\subsection{No-reference Image Quality Assessment}

Classical NR-IQA methods often rely on natural scene statistics (NSS) as a proxy for perceptual quality. BRISQUE~\cite{BRISQUE} regresses quality from spatial-domain NSS features of locally normalized luminance using a learned mapping, providing a distortion-generic yet hand-crafted baseline. NIQE~\cite{NIQE} removes the need for human opinion labels by measuring deviations from an NSS model of pristine natural images, making it a completely blind and opinion-unaware alternative. ILNIQE~\cite{ILNIQE} extends this idea by integrating multiple NSS cues and computing a patch-level distance to a learned pristine model, which often improves cross-dataset robustness. These NSS-based metrics are simple and interpretable, but their priors can misalign with enhancement outputs where naturalness is intentionally altered, such as contrast boosting or tone remapping, and where artifacts differ from legacy distortions. %

Modern NR-IQA increasingly relies on deep features and attention or Transformer backbones trained on large subjective datasets~\cite{duan2023attentive,duan2022confusing,duan2025finevq,min2024perceptual}. DBCNN~\cite{DBCNN} combines two CNN streams targeting synthetic and authentic distortions through bilinear pooling to better cover heterogeneous distortion regimes. HyperIQA~\cite{HyperIQA} uses a self-adaptive hyper-network to generate content-aware prediction parameters, improving generalization to images captured in the wild. MANIQA~\cite{yang2022maniqa} adopts a ViT-style backbone with multi-dimensional attention blocks to model global and local interactions for quality prediction. MUSIQ~\cite{MUSIQ} proposes a multi-scale Transformer that processes multiple resolutions to handle varying image sizes and capture quality cues at different granularities. TReS~\cite{TReS} further introduces learning signals beyond pointwise regression through relative ranking and self-consistency constraints, improving monotonicity and robustness to simple transformations. Recent work has also explored reducing dependence on large-scale MOS supervision, including label-efficient~\cite{LEAF} and label-free~\cite{ELIQ} IQA frameworks. While these models substantially advance generic NR-IQA, they are typically trained for broad degradations such as compression, blur, and noise, and may under-represent enhancement-specific failure modes and subjective criteria such as reasonable brightness or pleasing color. %

Datasets underpinning these advances range from controlled synthetic distortions to authentic images captured in the wild. The LIVE Challenge database, also known as CLIVE~\cite{ghadiyaram2016clive}, popularized authentically distorted images with large-scale crowdsourced ratings. KonIQ-10k~\cite{hosu2020koniq10k} expanded in-the-wild IQA to more than 10k images and enabled stronger deep models and cross-dataset evaluation. SPAQ~\cite{fang2020spaq} focuses on smartphone photography and provides rich attribute annotations such as noisiness and sharpness alongside MOS. PaQ-2-PiQ~\cite{ying2020paq2piq} offers large-scale picture-level and patch-level labels that connect local quality maps with global quality prediction. However, EIQA often differs from distorted-capture IQA. Enhancement outputs can be systematically biased by the enhancement algorithm and may shift image statistics in ways not represented by capture degradations, which limits direct transfer. 

\subsection{Image Enhancement Quality Assessment}

EIQA targets the perceptual quality of algorithmically enhanced images, where the best output may not coincide with a known reference and may even exceed the input image in perceived quality\cite{liu2025moa}. Early work also explored using IQA as an optimization signal. The no-reference model in~\cite{gu2019enhanced} was designed for enhanced images and used to guide iterative enhancement, illustrating the coupling between assessment and enhancement. In low-light enhancement, LIEQ~\cite{zhai2021lieqa} introduced a dedicated database together with a full-reference low-light enhancement index LIEQA that decomposes quality into luminance enhancement, color rendition, noise, and structure preservation, using multi-exposure fusion or HDR surrogates as references. Another line of work on low-light enhancement quality assessment constructed a subjective dataset and proposed a multi-feature reconciliation model that explicitly accounts for brightness, color, structure, and naturalness~\cite{nrlowlightqa}. These studies highlight that enhancement artifacts are multi-factor and are often not well explained by generic NR-IQA distortions. %

More recent EIQA studies emphasize large-scale subjective evaluation and tighter integration between evaluation and training. The Loop Game framework~\cite{chen2022loopgame} introduced the QUOTE-LOL dataset and iteratively alternated between enhancement and assessment to progressively improve results. RLIE~\cite{li2025rlie} further moves toward real-world low-light enhancement evaluation by constructing a dataset containing multiple enhanced versions of each scene and collecting pairwise judgments that are aggregated into global scores, together with an objective assessment model tailored to real-world low-light enhancement artifacts. 

\subsection{Algorithm Bias and Preference Modeling}

A growing body of work indicates that IQA models can be biased by distortion distributions, content priors, and dataset construction choices, which may lead to degraded performance under domain shift. Source-free unsupervised domain adaptation for BIQA~\cite{liu2022sfuda_biqa} addresses synthetic-to-authentic shift without access to source data by shaping prediction distributions with self-supervised objectives. DGQA~\cite{dgqa} further studies adaptation with distortion guidance and shows that increasing the number of synthetic distortions does not necessarily improve authentic generalization, while proposing mechanisms to bridge this gap. Continual and lifelong settings also reveal bias and drift. LIQA~\cite{liu2023liqa} studies BIQA under continuously changing distortion distributions and proposes a lifelong learning strategy to maintain performance over time. From a multimodal perspective, vision-language correspondence has been explored to introduce semantic priors into BIQA~\cite{zhang2023vlcorrespondence}, which may improve generalization but can also inherit biases from pretraining corpora. Collectively, these studies suggest that robust EIQA must explicitly address distribution shift and algorithm-dependent artifacts rather than assuming a stationary distortion universe. %

Preference modeling provides another perspective. Absolute MOS regression can be noisy, whereas pairwise comparisons are often more consistent for human observers. RankIQA~\cite{liu2017rankiqa} learns from relative rankings derived from known synthetic degradation orderings and transfers this supervision to absolute NR-IQA, demonstrating the data efficiency of ranking-based learning. PieAPP~\cite{prashnani2018pieapp} constructs a large-scale pairwise preference dataset for perceptual error and trains a model to predict human preferences, showing strong alignment with perceptual judgments in a full-reference setting. LPIPS~\cite{zhang2018lpips} learns a perceptual similarity metric directly from human judgments and performs metric learning in deep feature space to better match perceptual distances than classical metrics. When pairwise comparisons are collected, the Bradley--Terry model~\cite{bradley1952rank} provides a principled way to convert comparisons into global scores, as adopted in modern EIQA datasets such as RLIE~\cite{li2025rlie}. More recently, large pre-trained vision-language models have been explored for zero-shot perceptual assessment. CLIP-IQA~\cite{wang2023clipiqa} uses prompt pairing to convert CLIP similarity into a perceptual quality score without task-specific training, while ImageReward~\cite{xu2023imagereward} learns a reward model from expert comparisons to represent human preferences for text-to-image generation. Although effective, preference-based pipelines may still inherit bias from rater populations, prompt design, and the set of candidate algorithms being compared. These observations motivate our method to treat EIQA as a preference- and bias-aware problem rather than purely a score regression task, and to learn a perceptual space that remains stable across enhancement algorithms, scenes, and supervision formats.

\section{Method}
\label{chap:method}

\begin{figure*}
\centering
\includegraphics[width=0.9\linewidth]{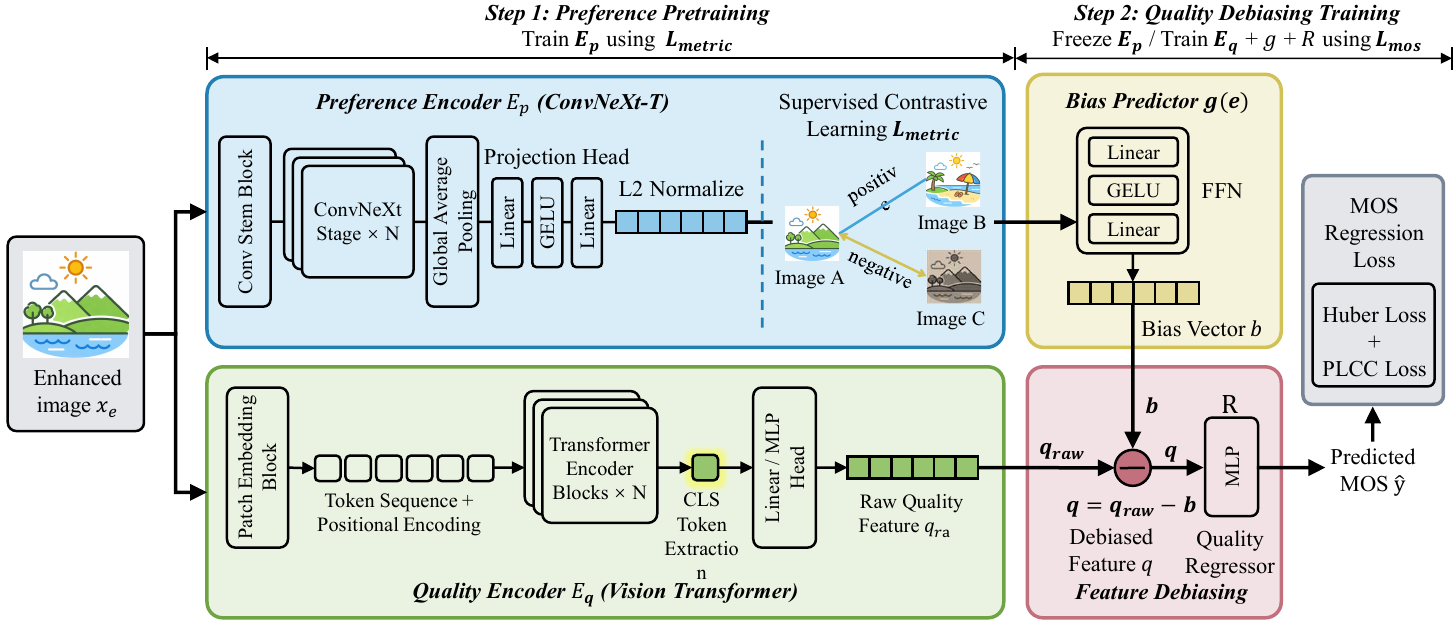}
\caption{Overall framework of the proposed method. First, a preference encoder is pretrained with supervised contrastive learning to capture enhancement-preference representations. Second, the learned preference cue is fed into a bias predictor to estimate the bias vector, which is then subtracted from the raw quality feature to obtain a debiased representation for MOS regression.}
\label{fig2}
\end{figure*}

\subsection{Problem Formulation and Framework Overview}

\subsubsection{Problem Formulation}
\label{sec:formulation}

We study no-reference enhancement image quality assessment (EIQA), where the goal is to predict a perceptual quality score for an enhanced image. 
Each training sample consists of a source raw image, its enhanced version, the enhancement algorithm label, and a subjective mean opinion score (MOS). 
Let $x_i^{r} \in \mathcal{X}$ denote the source raw image, $x_i^{e} \in \mathcal{X}$ denote the corresponding enhanced image, $a_i \in \{1,\dots,K\}$ denote the enhancement algorithm label, and $y_i \in \mathbb{R}$ denote the MOS. 
The training set is defined as
\begin{equation}
\mathcal{D}_{\text{train}} = \left\{(x_i^{r}, x_i^{e}, a_i, y_i)\right\}_{i=1}^{N}.
\end{equation}

Our target setting is no-reference EIQA. Therefore, at inference time, only the enhanced image is available, while the source raw image and the enhancement algorithm label are unknown. The goal is to learn a predictor
\begin{equation}
\hat{y} = f(x^{e}),
\end{equation}
which estimates perceptual quality from the enhanced image alone.

A key challenge in EIQA is that enhancement algorithms often introduce systematic, algorithm-dependent appearance variations. As a result, images generated by different algorithms may occupy distinct feature distributions even when they have similar perceptual quality. This may cause a quality predictor to rely on enhancement-specific fingerprints rather than intrinsic quality-related cues, leading to poor cross-algorithm generalization. To address this issue, we explicitly model enhancement preference and use it to remove algorithm-induced nuisance components from the quality representation.

\subsubsection{Framework Overview}
\label{sec:overview}

Our framework disentangles enhancement-specific preference from quality-relevant information. It consists of four components:
\begin{itemize}
    \item \textbf{Preference Encoder} $\mathcal{E}_{p}$, which extracts an enhancement-preference embedding $e$ from the enhanced image $x^{e}$;
    \item \textbf{Quality Encoder} $\mathcal{E}_{q}$, which extracts a raw quality representation $q_{\mathrm{raw}}$ from $x^{e}$;
    \item \textbf{Bias Predictor} $g(\cdot)$, which predicts an algorithm-induced bias vector from $e$;
    \item \textbf{Quality Regressor} $\mathcal{R}(\cdot)$, which predicts MOS from the debiased quality representation.
\end{itemize}
The schematic diagram is shown in Fig.~\ref{fig2}. Given an enhanced image $x^{e}$, the forward process is
\begin{align}
e &= \mathcal{E}_{p}(x^{e}), \label{eq:pref_embed}\\
q_{\mathrm{raw}} &= \mathcal{E}_{q}(x^{e}), \label{eq:qraw}\\
b &= g(e), \label{eq:bias_predict}\\
q &= q_{\mathrm{raw}} - b, \label{eq:debias_q}\\
\hat{y} &= \mathcal{R}(q). \label{eq:mos_predict}
\end{align}

Here, the preference encoder captures enhancement-style similarity, while the bias predictor maps the preference embedding to a nuisance component in the quality feature space. By subtracting this component from the raw quality representation, the resulting feature is encouraged to be more invariant to enhancement style and more aligned with perceptual quality.

\begin{figure*}
\centering
\includegraphics[width=0.99\linewidth]{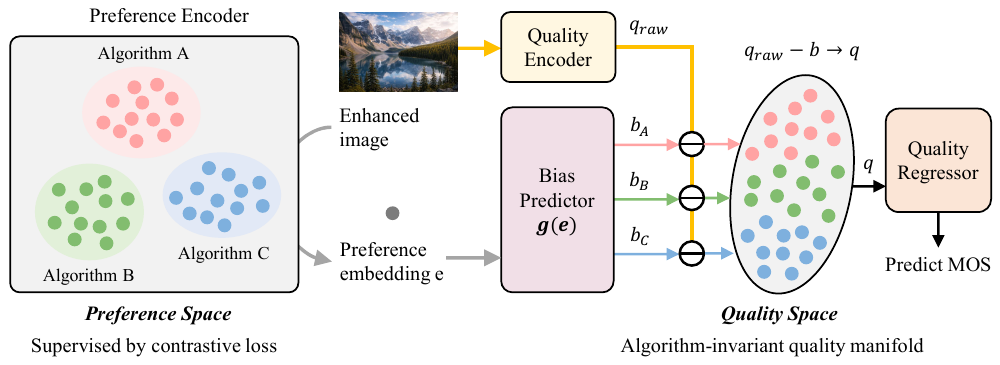}
\caption{Illustration of the proposed preference-guided debiasing method for enhancement image quality assessment. The model first learns a preference embedding from enhanced images to characterize algorithm-dependent enhancement styles, then predicts the corresponding bias term and removes it from the raw quality feature, yielding an algorithm-invariant quality representation for more reliable MOS prediction.}
\label{fig3}
\end{figure*}

\subsection{Preference Encoder}
\label{sec:metric}

\subsubsection{Motivation}
\label{subsec:pref_motivation}

Although enhancement algorithm labels are available during training, they are unavailable at inference time. Therefore, we do not formulate enhancement modeling as a closed-set classification problem. Instead, we learn a continuous embedding space in which enhanced images with similar algorithmic styles are close to each other, while those with different styles are separated. Compared with explicit algorithm classification, such a metric space provides a more flexible and transferable representation of enhancement preference.

\subsubsection{Supervised Contrastive Learning}
\label{subsec:supcon}

We implement $\mathcal{E}_{p}$ using ConvNeXt-T as the visual backbone, followed by a projection head that outputs a normalized embedding $e \in \mathbb{R}^{d}$. The preference encoder is trained using supervised contrastive learning. For a mini-batch $\mathcal{B}$, the positive set for sample $i$ is defined as
\begin{equation}
\mathcal{P}(i)=\{j \in \mathcal{B}\setminus\{i\}\mid a_j = a_i\}.
\end{equation}
The supervised contrastive loss is
\begin{equation}
\mathcal{L}_{\mathrm{metric}}
=
\sum_{i \in \mathcal{B}}
\frac{-1}{|\mathcal{P}(i)|}
\sum_{j \in \mathcal{P}(i)}
\log
\frac{\exp(\mathrm{sim}(e_i,e_j)/\tau)}
{\sum_{k \in \mathcal{B}\setminus\{i\}} \exp(\mathrm{sim}(e_i,e_k)/\tau)},
\label{eq:supcon}
\end{equation}
where $\mathrm{sim}(\cdot,\cdot)$ denotes cosine similarity and $\tau$ is a temperature parameter.

This objective encourages the embedding space to preserve enhancement-style similarity without imposing rigid decision boundaries as in closed-set classification.

\subsubsection{Content-Controlled Sampling}
\label{subsec:sampling}

A major challenge in learning enhancement preference is content leakage, namely, the encoder relying on scene semantics instead of enhancement characteristics. To alleviate this issue, we construct mini-batches in a content-aware manner. Specifically, for a raw image with multiple enhanced versions, we explicitly sample images sharing the same source content but generated by different enhancement algorithms into the same mini-batch.

Under the supervised contrastive loss in Eq.~\eqref{eq:supcon}, these samples act as hard negatives: although they are semantically similar, they belong to different enhancement styles. This batch construction forces the preference encoder to focus on algorithm-dependent appearance characteristics rather than image content. Note that the raw image $x^{r}$ is used only for content-controlled sampling during training and is not required as model input.

\subsection{Preference-Guided Debiasing for Quality Representation}
\label{sec:debias}

In EIQA, quality features extracted from enhanced images often entangle perceptual quality with enhancement-specific style characteristics. Different algorithms may introduce distinct artifacts, tonal shifts, or texture patterns, causing the model to encode algorithm-dependent information rather than intrinsic quality cues. To address this issue, we use the preference embedding to guide quality feature debiasing. The schematic diagram is shown in Fig.~\ref{fig3}. 

\subsubsection{Bias Subtraction}
\label{subsec:bias_subtract}

We model the enhancement-induced nuisance component as an additive bias in the quality feature space. This can be viewed as a first-order approximation of the algorithm-induced feature shift, which provides a lightweight and interpretable mechanism for suppressing dominant style-related variation.

Specifically, the raw quality representation $q_{\mathrm{raw}}$ is extracted by a quality encoder $\mathcal{E}_{q}$ implemented with a ViT backbone. A lightweight bias predictor $g(\cdot)$, implemented as a two-layer MLP with a GELU activation, maps the preference embedding to a bias vector
\begin{equation}
b = g(e), \qquad b \in \mathbb{R}^{D},
\end{equation}
where $D$ is the dimensionality of the quality feature. The debiased representation is then obtained by
\begin{equation}
q = q_{\mathrm{raw}} - b.
\end{equation}

By subtracting the predicted nuisance component, the resulting feature $q$ is encouraged to suppress algorithm-specific bias and retain perceptual quality information that is more transferable across enhancement styles.

\subsubsection{MOS Regression}
\label{subsec:mos_loss}

The final MOS prediction is obtained from the debiased feature:
\begin{equation}
\hat{y} = \mathcal{R}(q),
\end{equation}
where $\mathcal{R}(\cdot)$ denotes the quality regressor. We optimize the prediction branch using a combination of Huber loss and PLCC loss:
\begin{equation}
\mathcal{L}_{\mathrm{mos}}
=
\mathcal{L}_{\mathrm{Huber}}
+
\lambda_{\mathrm{plcc}} \mathcal{L}_{\mathrm{plcc}},
\label{eq:mosloss}
\end{equation}
where
\begin{equation}
\mathcal{L}_{\mathrm{Huber}}
=
\frac{1}{|\mathcal{B}|}
\sum_{i \in \mathcal{B}}
\rho(\hat{y}_i - y_i),
\end{equation}
and
\begin{equation}
\mathcal{L}_{\mathrm{plcc}} = 1 - \mathrm{PLCC}(\hat{\mathbf{y}}, \mathbf{y}).
\end{equation}
Here, $\rho(\cdot)$ denotes the Huber penalty, and $\mathrm{PLCC}(\hat{\mathbf{y}}, \mathbf{y})$ measures the linear correlation between predicted scores and ground-truth MOS values within a mini-batch. The Huber term improves point-wise regression accuracy, while the PLCC term encourages global linear alignment with MOS.

\subsection{Optimization Strategy}
\label{sec:overall}

To separate the functional roles of preference modeling and quality prediction, we adopt a two-stage training strategy.

For preference pretraining, we first train the preference encoder $\mathcal{E}_{p}$ using the metric loss in Eq.~\eqref{eq:supcon}.
The goal of this stage is to establish a stable enhancement-preference embedding space that captures enhancement-style similarity.

For quality debiasing and regression, we freeze $\mathcal{E}_{p}$ and train the quality encoder $\mathcal{E}_{q}$, the bias predictor $g(\cdot)$, and the regressor $\mathcal{R}$ using the MOS regression loss in Eq.~\eqref{eq:mosloss}.
During this stage, the pretrained preference embedding $e=\mathcal{E}_{p}(x^{e})$ serves as a fixed conditioning signal for bias prediction. In this way, the preference branch remains dedicated to enhancement-style encoding, while the quality branch learns perceptual quality prediction from debiased features.

This stage-wise design improves optimization stability and reduces interference between the two objectives.

\subsubsection{Inference}
\label{subsec:inference}

At inference time, only the enhanced image is required. Given $x^{e}$, we compute
\begin{equation}
e=\mathcal{E}_{p}(x^{e}), \qquad
q_{\mathrm{raw}}=\mathcal{E}_{q}(x^{e}), \qquad
q=q_{\mathrm{raw}}-g(e), \qquad
\hat{y}=\mathcal{R}(q).
\end{equation}
Neither the raw image nor the enhancement algorithm label is needed during testing. Therefore, the proposed framework is fully compatible with the no-reference EIQA setting.


\section{Experiments}
\subsection{Datasets and Evaluation Protocols}
\label{sec:exp_settings}

We conduct experiments on two public enhancement image quality assessment benchmarks, namely RLIE and SQUARE-LOL. RLIE contains 1,540 enhanced low-light images generated from 154 real-world scenes. Following its official protocol, we adopt \emph{5-fold cross-validation} with environment-aware splitting for evaluation. SQUARE-LOL contains 2,900 enhanced images derived from 290 natural images using 10 enhancement algorithms. Following prior EIQA studies, we use its standard evaluation protocol for benchmark comparison and ablation analysis.

In addition to the standard setting, we further construct an \emph{algorithm-disjoint} protocol on SQUARE-LOL to evaluate cross-algorithm generalization. Specifically, images generated by a subset of enhancement algorithms are used for training, while those generated by unseen algorithms are reserved for testing. Unless otherwise specified, 8 algorithms are used for training and the remaining 2 for testing.

We report Spearman rank-order correlation coefficient (SRCC) and Pearson linear correlation coefficient (PLCC) as the primary evaluation metrics. For the main comparison tables, we additionally provide the Kendall rank-order correlation coefficient (KRCC). Higher values indicate better agreement with subjective human judgments.

\subsection{Implementation Details}
\label{sec:impl_details}

Our model is implemented in PyTorch. The preference encoder is built on ConvNeXt-T, while the quality encoder adopts a ViT backbone. During training, input images are randomly cropped to $224\times224$, with standard data augmentation including random rotation and horizontal flipping. The model is optimized using Adam with an initial learning rate of $1\times10^{-4}$, a batch size of $32$, and a total of $30$ training epochs.

Following the two-stage optimization strategy described in Sec.~\ref{sec:overall}, we first pretrain the preference encoder using the supervised contrastive objective, and then freeze it to train the quality encoder, bias predictor, and quality regressor. Unless otherwise specified, all competing variants in the ablation study use the same backbone settings, training schedule, and evaluation protocol as the full model.

\begin{table}[t]
\centering
\setlength{\tabcolsep}{10pt}
\renewcommand{\arraystretch}{1.15}
\caption{Comparison with state-of-the-art no-reference IQA methods on the RLIE dataset.}
\label{tab_rlie}
\resizebox{0.49\textwidth}{!}{
\begin{tabular}{lcccc}
\toprule
Methods & Type & SRCC & PLCC & KRCC \\
\midrule
NIQE~\cite{NIQE}               & Handcrafted   & 0.0200 & 0.0040 & 0.0240 \\
BRISQUE~\cite{BRISQUE}         & Handcrafted   & 0.0302 & 0.0463 & 0.0271 \\
ILNIQE~\cite{ILNIQE}           & Handcrafted   & 0.0956 & 0.0972 & 0.0565 \\
IS~\cite{IS}                   & Handcrafted   & 0.0869 & 0.0898 & 0.0483 \\
DBCNN~\cite{DBCNN}             & Deep-learning & 0.4631 & 0.5131 & 0.3640 \\
HyperIQA~\cite{HyperIQA}       & Deep-learning & 0.6158 & 0.6702 & 0.4828 \\
MUSIQ~\cite{MUSIQ}             & Deep-learning & 0.6284 & 0.6402 & 0.5023 \\
UNIQUE~\cite{UNIQUE}           & Deep-learning & 0.6429 & 0.6607 & 0.5270 \\
TReS~\cite{TReS}               & Deep-learning & 0.5746 & 0.5975 & 0.4614 \\
StairIQA~\cite{StairIQA}       & Deep-learning & 0.4665 & 0.5213 & 0.3286 \\
SGRNet~\cite{SGRNet}           & Deep-learning & 0.5604 & 0.5728 & 0.4452 \\
AGAIQA~\cite{AGAIQA}           & Deep-learning & 0.5785 & 0.5729 & 0.4564 \\
IACA~\cite{IACA}               & Deep-learning & 0.6067 & 0.6705 & 0.4807 \\
MIIHDP~\cite{MIIHDP}           & Deep-learning & 0.6682 & 0.6890 & 0.5378 \\
\textbf{Ours}                  & Deep-learning & \textbf{0.6923} & \textbf{0.7137} & \textbf{0.5516} \\
\bottomrule
\end{tabular}
}
\end{table}

\subsection{Comparison with State-of-the-Art Methods}
\label{sec:sota_comparison}

We compare the proposed method with $14$ representative no-reference IQA models on RLIE and SQUARE-LOL, including four hand-crafted methods and ten deep-learning-based approaches. For fair comparison, all methods are evaluated under the corresponding standard protocol of each dataset.

As shown in Tables~\ref{tab_rlie} and \ref{tab_squarelol}, the proposed method achieves the best performance on both datasets. On RLIE, our method attains SRCC/PLCC of $0.6923$/$0.7137$, surpassing the strongest competing method by $0.0241$ and $0.0247$, respectively. On SQUARE-LOL, our method further achieves SRCC/PLCC of $0.8726$/$0.8913$, improving over the best existing result by $0.0421$ and $0.0429$, respectively.

Several observations can be drawn from these results. First, traditional hand-crafted BIQA methods perform poorly on both EIQA datasets, suggesting that natural scene statistic features are insufficient to characterize the complex appearance changes introduced by image enhancement algorithms. Second, although deep-learning-based IQA models perform substantially better, their predictions remain vulnerable to enhancement-style variation. Third, our method consistently outperforms all competing approaches, which verifies the effectiveness of explicitly modeling enhancement preference and removing the corresponding nuisance component from the quality representation.

\begin{table}[t]
\centering
\setlength{\tabcolsep}{10pt}
\renewcommand{\arraystretch}{1.15}
\caption{Comparison with state-of-the-art no-reference IQA methods on the SQUARE-LOL dataset.}
\label{tab_squarelol}
\resizebox{0.49\textwidth}{!}{
\begin{tabular}{lcccc}
\toprule
Methods & Type & SRCC & PLCC & KRCC \\
\midrule
NIQE~\cite{NIQE}               & Handcrafted   & 0.1378 & 0.1572 & 0.0550 \\
BRISQUE~\cite{BRISQUE}         & Handcrafted   & 0.1532 & 0.1813 & 0.0593 \\
ILNIQE~\cite{ILNIQE}           & Handcrafted   & 0.1807 & 0.1975 & 0.0610 \\
IS~\cite{IS}                   & Handcrafted   & 0.1631 & 0.1760 & 0.0574 \\
DBCNN~\cite{DBCNN}             & Deep-learning & 0.7934 & 0.8056 & 0.6373 \\
HyperIQA~\cite{HyperIQA}       & Deep-learning & 0.7419 & 0.7676 & 0.5769 \\
MUSIQ~\cite{MUSIQ}             & Deep-learning & 0.7729 & 0.7942 & 0.6290 \\
UNIQUE~\cite{UNIQUE}           & Deep-learning & 0.7766 & 0.7899 & 0.6215 \\
TReS~\cite{TReS}               & Deep-learning & 0.7803 & 0.8002 & 0.6286 \\
StairIQA~\cite{StairIQA}       & Deep-learning & 0.6932 & 0.7226 & 0.5255 \\
SGRNet~\cite{SGRNet}           & Deep-learning & 0.7223 & 0.7356 & 0.5741 \\
AGAIQA~\cite{AGAIQA}           & Deep-learning & 0.6824 & 0.7020 & 0.5213 \\
IACA~\cite{IACA}               & Deep-learning & 0.8305 & 0.8484 & 0.6530 \\
MIIHDP~\cite{MIIHDP}           & Deep-learning & 0.8246 & 0.8389 & 0.6442 \\
\textbf{Ours}                  & Deep-learning & \textbf{0.8726} & \textbf{0.8913} & \textbf{0.6904} \\
\bottomrule
\end{tabular}
}
\end{table}

\subsection{Cross-Algorithm Generalization Evaluation}
\label{sec:cross_algorithm}

To further evaluate robustness to algorithm-induced distribution shifts, we conduct a cross-algorithm generalization experiment on the SQUARE-LOL dataset. Different from the standard protocol, this setting adopts an \emph{algorithm-disjoint} split: images generated by a subset of enhancement algorithms are used for training, while images generated by previously unseen algorithms are used for testing. In our experiments, the dataset contains 10 enhancement algorithms in total, among which 8 algorithms are used for training, and the remaining 2 are reserved for testing.

This protocol is designed to directly evaluate generalization across enhancement algorithms. Therefore, the train/test split is algorithm-disjoint. Compared with the standard setting, it provides a more direct test of whether a model relies on algorithm-specific fingerprints or learns a more transferable quality representation.

Table~\ref{tab:cross_algorithm} reports the results of representative no-reference IQA methods under both the standard protocol and the unseen-algorithm protocol. We further report the performance drop from the standard setting to the unseen-algorithm setting. As can be seen, all competing methods suffer clear performance degradation when evaluated on unseen enhancement algorithms, confirming that algorithm-induced distribution shift is a major challenge in EIQA.

In contrast, the proposed method shows the strongest robustness. Specifically, our method achieves the best unseen-algorithm performance, reaching $0.8622$ SRCC and $0.8804$ PLCC. More importantly, it yields the smallest performance drop, with only $0.0104$ in SRCC and $0.0109$ in PLCC. By comparison, the drops of existing methods are substantially larger, such as $0.0689$/$0.0750$ for IACA and $0.0697$/$0.0731$ for TReS in SRCC/PLCC. These results indicate that existing BIQA models are still sensitive to enhancement-specific bias, while the proposed preference-guided debiasing strategy effectively suppresses such nuisance information and improves transferability to unseen enhancement algorithms.

Overall, this experiment provides direct evidence that the proposed method not only improves performance under the standard setting but also generalizes better when the testing images are produced by unseen enhancement algorithms.

\begin{table}[t]
\centering
\caption{Cross-algorithm generalization evaluation on the SQUARE-LOL dataset under the algorithm-disjoint protocol. ``Standard'' denotes the result under the default evaluation protocol, ``Unseen'' denotes the result on unseen enhancement algorithms, and ``Drop'' is the performance decrease from Standard to Unseen. The best results are highlighted in bold.}
\label{tab:cross_algorithm}
\resizebox{0.49\textwidth}{!}{
\begin{tabular}{lcccccc}
\toprule
& \multicolumn{3}{c}{SRCC $\uparrow$} & \multicolumn{3}{c}{PLCC $\uparrow$} \\
\cmidrule(lr){2-4} \cmidrule(lr){5-7}
Methods & Standard & Unseen & Drop $\downarrow$ & Standard & Unseen & Drop $\downarrow$ \\
\midrule
DBCNN~\cite{DBCNN}         & 0.7934 & 0.7406 & 0.0528 & 0.8056 & 0.7626 & 0.0430 \\
HyperIQA~\cite{HyperIQA}   & 0.7419 & 0.6911 & 0.0508 & 0.7676 & 0.7132 & 0.0544 \\
MUSIQ~\cite{MUSIQ}         & 0.7729 & 0.7424 & 0.0305 & 0.7942 & 0.7644 & 0.0298 \\
UNIQUE~\cite{UNIQUE}       & 0.7766 & 0.7309 & 0.0457 & 0.7899 & 0.7597 & 0.0302 \\
TReS~\cite{TReS}           & 0.7803 & 0.7106 & 0.0697 & 0.8002 & 0.7271 & 0.0731 \\
StairIQA~\cite{StairIQA}   & 0.6932 & 0.6449 & 0.0483 & 0.7226 & 0.6607 & 0.0619 \\
SGRNet~\cite{SGRNet}       & 0.7223 & 0.6678 & 0.0545 & 0.7356 & 0.6842 & 0.0514 \\
AGAIQA~\cite{AGAIQA}       & 0.6824 & 0.6458 & 0.0366 & 0.7020 & 0.6676 & 0.0344 \\
IACA~\cite{IACA}           & 0.8305 & 0.7616 & 0.0689 & 0.8484 & 0.7734 & 0.0750 \\
\textbf{Ours}              & \textbf{0.8726} & \textbf{0.8622} & \textbf{0.0104} & \textbf{0.8913} & \textbf{0.8804} & \textbf{0.0109} \\
\bottomrule
\end{tabular}
}
\end{table}

\subsection{Ablation Studies}
\label{sec:ablation}

Unless otherwise specified, all ablation experiments are conducted on the SQUARE-LOL dataset under the standard evaluation protocol. For fair comparison, each variant keeps the same backbone architecture, optimization setting, and evaluation protocol as the full model, while only modifying the component under investigation. We report SRCC, PLCC and KRCC in this part.

\subsubsection{Effect of Preference-Guided Debiasing}
\label{subsec:ablation_debiasing}

We evaluate the effect of introducing preference information and the proposed debiasing strategy. Three variants are compared: \textbf{w/o Preference}, which removes the preference branch and predicts MOS directly from the quality feature; \textbf{Preference Concat}, which incorporates the preference embedding by concatenating it with the quality feature for regression; and \textbf{Preference-Guided Debiasing}, i.e., the full model, which predicts an enhancement-related bias vector from the preference embedding and subtracts it from the raw quality feature.

As shown in Table~\ref{tab:ablation_debiasing}, introducing preference information is beneficial for EIQA. Compared with \textbf{w/o Preference}, \textbf{Preference Concat} improves SRCC/PLCC/KRCC from 0.8557/0.8689/0.6802 to 0.8616/0.8773/0.6834, indicating that the preference branch provides complementary information for quality prediction. Furthermore, the proposed \textbf{Preference-Guided Debiasing} achieves the best performance of 0.8726/0.8913/0.6904, outperforming \textbf{w/o Preference} by 0.0169/0.0224/0.0102 and \textbf{Preference Concat} by 0.0110/0.0140/0.0070 in SRCC/PLCC/KRCC, respectively. These results suggest that the performance gain is not merely due to introducing an additional preference branch, but mainly comes from explicitly modeling and removing enhancement-related bias from the quality representation.

\begin{table}[t]
\centering
\caption{Effect of preference-guided debiasing on the SQUARE-LOL dataset. The best results are highlighted in bold.}
\label{tab:ablation_debiasing}
\resizebox{\linewidth}{!}{
\begin{tabular}{lccc}
\toprule
Method & SRCC $\uparrow$ & PLCC $\uparrow$ & KRCC $\uparrow$ \\
\midrule
w/o Preference & 0.8557 & 0.8689 & 0.6802 \\
+ Preference Concat & 0.8616 & 0.8773 & 0.6834 \\
+ Preference-Guided Debiasing & \textbf{0.8726} & \textbf{0.8913} & \textbf{0.6904} \\
\bottomrule
\end{tabular}
}
\end{table}

\subsubsection{Effect of Preference Representation Learning}
\label{subsec:ablation_pref_learning}

We next study how the preference representation should be learned. Here, the compared variants share the same debiasing framework, while only the preference learning objective is changed. We compare three settings: \textbf{w/o Preference}, which removes the preference branch; \textbf{Cls-based Preference}, which learns preference features by enhancement algorithm classification; and \textbf{SupCon-based Preference}, which is our final design.

Table~\ref{tab:ablation_pref_learning} shows that removing the preference branch gives the weakest result, with SRCC/PLCC of 0.8557/0.8689. Using classification-based preference learning improves the performance to 0.8701/0.8865, indicating that enhancement algorithm supervision is useful. Our supervised contrastive design achieves the best performance, reaching 0.8726/0.8913. This suggests that a contrastive preference space is more effective than a closed-set classification space for modeling enhancement style and supporting downstream debiasing.

\begin{table}[t]
\centering
\caption{Effect of preference representation learning on the SQUARE-LOL dataset. The best results are highlighted in bold.}
\label{tab:ablation_pref_learning}
\resizebox{\linewidth}{!}{
\begin{tabular}{lccc}
\toprule
Method & SRCC $\uparrow$ & PLCC $\uparrow$ & KRCC $\uparrow$ \\
\midrule
w/o Preference & 0.8557 & 0.8689 & 0.6802 \\
Cls-based Preference & 0.8701 & 0.8865 & 0.6892 \\
SupCon-based Preference (Ours) & \textbf{0.8726} & \textbf{0.8913} & \textbf{0.6904} \\
\bottomrule
\end{tabular}
}
\end{table}

\subsubsection{Effect of Two-Stage Training}
\label{subsec:ablation_twostage}

We further evaluate the proposed two-stage training strategy. Three settings are compared: \textbf{Joint Training}, which optimizes all modules together from scratch; \textbf{Two-Stage w/o Freezing}, which pretrains the preference encoder and then continues joint optimization without freezing it; and \textbf{Two-Stage w/ Freezing}, which is our final design.

As shown in Table~\ref{tab:ablation_twostage}, joint training yields SRCC/PLCC of 0.8510/0.8645. Stage-wise optimization without freezing already improves the performance to 0.8673/0.8792, showing the benefit of learning a structured preference space before regression. Freezing the preference encoder in the second stage further improves the performance to 0.8726/0.8913. These results indicate that direct joint optimization introduces interference between preference learning and MOS regression, while the proposed two-stage strategy preserves a more stable preference space and provides more reliable guidance for debiasing.

\begin{table}[t]
\centering
\caption{Effect of different training strategies on the SQUARE-LOL dataset. The best results are highlighted in bold.}
\label{tab:ablation_twostage}
\resizebox{\linewidth}{!}{
\begin{tabular}{lccc}
\toprule
Training Strategy & SRCC $\uparrow$ & PLCC $\uparrow$ & KRCC $\uparrow$ \\
\midrule
Joint Training & 0.8510 & 0.8645 & 0.6809 \\
Two-Stage w/o Freezing & 0.8673 & 0.8792 & 0.6864 \\
Two-Stage w/ Freezing (Ours) & \textbf{0.8726} & \textbf{0.8913} & \textbf{0.6904} \\
\bottomrule
\end{tabular}
}
\end{table}

\begin{table}[t]
\centering
\caption{Effect of different sampling strategies for preference learning on the SQUARE-LOL dataset. The best results are highlighted in bold.}
\label{tab:ablation_sampling}
\resizebox{\linewidth}{!}{
\begin{tabular}{lccc}
\toprule
Sampling Strategy & SRCC $\uparrow$ & PLCC $\uparrow$ & KRCC $\uparrow$ \\
\midrule
Random Sampling & 0.8527 & 0.8656 & 0.6868 \\
Algorithm-Balanced Sampling & 0.8578 & 0.8690 & 0.6863 \\
Content-Controlled Sampling (Ours) & \textbf{0.8726} & \textbf{0.8913} & \textbf{0.6904} \\
\bottomrule
\end{tabular}
}
\end{table}

\subsubsection{Effect of Content-Controlled Sampling}
\label{subsec:ablation_sampling}

We finally evaluate the content-controlled sampling strategy used in preference learning. The goal of this design is to reduce content leakage, \emph{i.e.}, preventing the preference encoder from relying on scene semantics rather than enhancement-specific characteristics.

We compare three sampling strategies: \textbf{Random Sampling}, which forms mini-batches without considering algorithm balance or source-content relations; \textbf{Algorithm-Balanced Sampling}, which enforces a more balanced distribution of enhancement algorithms within each mini-batch; and \textbf{Content-Controlled Sampling}, which further introduces samples sharing the same source content but generated by different enhancement algorithms as hard negatives.

As shown in Table~\ref{tab:ablation_sampling}, random sampling achieves SRCC/PLCC of 0.8527/0.8656. Algorithm-balanced sampling slightly improves the performance to 0.8578/0.8690, indicating that a more balanced exposure to enhancement styles is beneficial. By further incorporating same-content, different-algorithm hard negatives, the proposed content-controlled sampling strategy achieves the best performance, reaching 0.8726/0.8913. Compared with random sampling, it brings gains of 0.0199 in SRCC and 0.0257 in PLCC. This verifies that explicitly controlling content interference helps the encoder focus more on enhancement-specific style cues and leads to better downstream quality prediction.

\subsection{Visualization and Analysis}

\begin{figure*}
\centering
\includegraphics[width=0.85\linewidth]{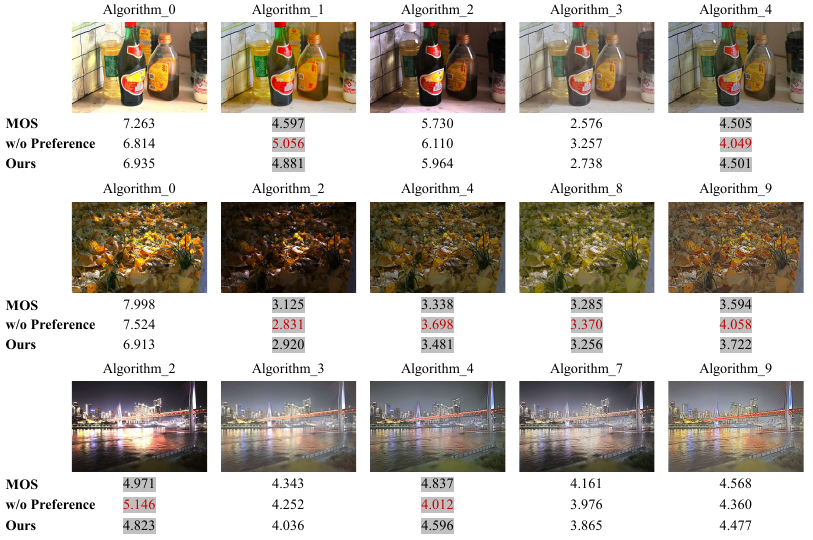}
\caption{Qualitative comparison on images enhanced by different algorithms.
Images in each row share the same source content but exhibit different enhancement styles.
Although their MOS values are relatively similar, the model without preference guidance produces inconsistent predictions,
while the proposed method yields predictions that are more stable and closer to the ground-truth MOS.
Red numbers indicate relatively large prediction errors.}
\label{fig4}
\end{figure*}

Figure~\ref{fig4} presents several qualitative examples where the same source image is enhanced by different algorithms.
Although these enhanced images exhibit noticeably different visual styles, their perceptual quality is often comparable according to the MOS.

However, the model without preference guidance tends to produce inconsistent predictions across different enhancement algorithms.
For example, in the first and second rows, the model significantly overestimates the quality of several enhanced images while underestimating others with similar perceptual quality.
This indicates that the raw quality representation is influenced by algorithm-specific appearance characteristics rather than intrinsic quality cues.

In contrast, the proposed method produces predictions that are closer to the MOS and more stable across different enhancement styles.
These results provide intuitive evidence that the proposed preference-guided debiasing mechanism helps suppress enhancement-specific bias and improves robustness across enhancement algorithms.

\section{Conclusion}

This paper presented a preference-guided debiasing framework for no-reference enhancement image quality assessment. Motivated by the observation that enhancement algorithms introduce algorithm-dependent style bias, we explicitly modeled enhancement-style preference and used it to remove nuisance components from quality representations. The proposed two-stage framework enables the model to focus on algorithm-invariant perceptual quality cues rather than enhancement-specific visual fingerprints. Experimental results showed that our method achieves robust quality prediction and improved cross-algorithm generalization on public EIQA benchmarks. These results suggest that explicitly disentangling enhancement style from perceptual quality is a promising direction for building more generalizable EIQA models.

\bibliographystyle{IEEEtran} 

\bibliography{arxiv}



\end{document}